\documentclass[conference]{IEEEtran}
\usepackage{newtxtext}

\usepackage{algorithmicx}
\usepackage{algpseudocode}

\usepackage{cite}
\usepackage{amsmath,amssymb,amsfonts}
\algrenewcommand\algorithmiccomment[1]{\hfill\textit{// #1}}
\usepackage{graphicx}
\usepackage{textcomp}
\usepackage{xcolor}
\usepackage{algorithm}
\usepackage{tabularx}
\usepackage{relsize}
\usepackage{hyperref}
\usepackage{url}
\def\BibTeX{{\rm B\kern-.05em{\sc i\kern-.025em b}\kern-.08em
    T\kern-.1667em\lower.7ex\hbox{E}\kern-.125emX}}

\IEEEspecialpapernotice{Accepted in The IEEE International Workshop on Large Language Models in Finance, Preprint Copy\\ Dec 8-11, Macau, China, 2025}
\begin{document}

\title{	SQuARE: Structured Query \& Adaptive Retrieval Engine For Tabular Formats
}

\author{\IEEEauthorblockN{Chinmay Gondhalekar}
\IEEEauthorblockA{\textit{Ratings Data Science} \\
\textit{S\&P Global}\\
New York, United States of America \\
chinmay.gondhalekar@spglobal.com}
\and
\IEEEauthorblockN{Urjitkumar Patel}
\IEEEauthorblockA{\textit{Ratings Data Science} \\
\textit{S\&P Global}\\
New York, United States of America \\
urjitkumar.patel@spglobal.com}
\and
\IEEEauthorblockN{Fang-Chun Yeh}
\IEEEauthorblockA{\textit{Ratings Data Science} \\
\textit{S\&P Global}\\
New York, United States of America \\
jessie.yeh@spglobal.com}
}

\maketitle

\begin{abstract}

Accurate question answering over real spreadsheets remains difficult due to multirow headers, merged cells, and unit annotations that disrupt naive chunking, while rigid SQL views fail on files lacking consistent schemas. We present \emph{SQuARE}, a hybrid retrieval framework with sheet-level, complexity-aware routing. It computes a continuous score based on header depth and merge density, then routes queries either through structure-preserving chunk retrieval or SQL over an automatically constructed relational representation. A lightweight agent supervises retrieval, refinement, or combination of results across both paths when confidence is low. This design maintains header hierarchies, time labels, and units, ensuring that returned values are faithful to the original cells and straightforward to verify.

Evaluated on multi-header corporate balance sheets, a heavily merged World Bank workbook, and diverse public datasets, SQuARE consistently surpasses single-strategy baselines and ChatGPT-4o on both retrieval precision and end-to-end answer accuracy while keeping latency predictable. By decoupling retrieval from model choice, the system is compatible with emerging tabular foundation models and offers a practical bridge toward a more robust table understanding.

\end{abstract}

\begin{IEEEkeywords}
Retrieval-Augmented Generation (RAG), Large Language Models, Natural Language Processing, Financial QA, Structured Data Q\&A, Deep Learning
\end{IEEEkeywords}

\section{Introduction}
Spreadsheets constitute the predominant medium for quantitative analysis across numerous disciplines, particularly in the field of finance. Yet, question answering (QA) over real world tabular format files often fails for two reasons. First, naive text chunking loses structure: multi-row headers and merged cells carry meaning (for example, a subtotal versus a grand total, unit lines, fiscal versus calendar years) that fixed windows split apart. Second, pure SQL assumes a tidy schema and stable column semantics; it breaks on header nesting, irregular unit rows, and ad hoc formatting. SQuARE detects spreadsheet structure and routes each query to the safer path: structure-preserving chunks when header paths or unit lines matter, or constrained SQL when filters/aggregations dominate. This avoids one size fits all failures and returns answers with exact cells/rows as auditable evidence.

Consider a corporate balance sheet with three header rows and a line that reads ``USD (millions)'' between the headers and the values. The answer to ``Total equity attributable to the parent in FY2023?'' depends on both the full header path leading to the target row and the unit line. Fixed-size chunks often separate those elements. An always-SQL approach can map to the wrong column or silently ignore the unit row. By contrast, a flat public dataset (for example, GDP growth by country and year) is well served by exact filters and aggregations, where SQL is both faster and safer.

We take a simple stance, do not treat all tables the same. We introduce \emph{SQuARE} (Structured Query and Adaptive Retrieval Engine), a framework with a set of decision functions and a routing mechanism. SQuARE scores a sheet’s structural complexity (header depth and merged cells) and routes each query either to structure-preserving chunk retrieval or to SQL over an automatically constructed relational view. A lightweight agent monitors confidence and, when needed, refines or merges evidence from both paths. The system surfaces the specific rows and cells used to answer, keeping header context and units intact when they matter, and using deterministic queries when they are helpful.

SQuARE constitutes a modular framework that facilitates the interchangeable use of embeddings, vector stores, and databases while maintaining an invariant control flow. The framework prioritizes numerical precision and verifiable provenance.

Our contributions are:
\begin{itemize}
  \item \textbf{Complexity metric for routing} We establish a metric that evaluates both header depth and the number of merged cells. By employing a single threshold, this metric effectively distinguishes sheets categorized as \emph{Multi-Header} from those categorized \emph{as Flat}, thereby activating minimal backend processing. 

  \item \textbf{Structure-preserving segmentation with semantic indexing} For intricate sheets, segmentation occurs at header boundaries, and metadata is incorporated to capture header paths, temporal labels, and units of measure. Rather than indexing raw cells, we utilize concise descriptions for embeddings.

  \item \textbf{Schema aware SQL generation with guardrails} For flat tables, we deduce a cleansed schema and produce constrained SQL through a concise refinement process. 

  \item \textbf{Agentic routing with confidence-aware fallback} : a lightweight agent selects \texttt{chunks} or \texttt{sql} based on complexity and query cues. On low confidence, it switches modes or merges contexts, summarizes to fit the budget if needed, and returns the exact rows/blocks used.
\end{itemize}

We evaluate across datasets with varying difficulty tiers. Under identical inputs, SQuARE outperforms single-strategy baselines and OpenAI’s ChatGPT-4o (tested via the public ChatGPT interface, default settings) \cite{gpt4o}; complete results are reported in the Evaluation section.

\section{Related Work}

Recent advances in retrieval-augmented generation (RAG) have begun to address the challenge of querying complex tables. For example, TabRAG \cite{xu2024tabrag} introduces a RAG pipeline that retrieves entire table \emph{images} from a large collection and feeds them to a large multimodal model (LMM) to generate answers. This approach avoids naively flattening tables into text, leveraging visual layout but still relies on purely semantic retrieval. In another line of work, TableRAG \cite{yu2025tableragretrievalaugmentedgeneration} specifically targets heterogeneous documents containing both text and tables. The authors observe that the common practice of linearizing tables and fixed-size chunking can ``disrupt the intrinsic tabular structure, lead to information loss,’’ and break multi-hop reasoning. TableRAG builds an offline relational database and performs \emph{mixed retrieval}: user queries are decomposed into sub-queries that are either answered by text retrieval or by generating and executing SQL on the table database \cite{yu2025tableragretrievalaugmentedgeneration, yu2025tablerag_extended}. This preserves table integrity and improves global reasoning; however, many implementations still fall back to some form of flattening and therefore suffer from structural information loss.

Other pipelines have proposed similar hybrid strategies. For instance, Khanna et al. \cite{khanna2024finetuning} demonstrates that fine-tuning embedding models specifically for tabular RAG greatly improves retrieval precision in financial datasets. Industry practitioners have also emphasized ``structured RAG’’ or SQL-augmented RAG: systems that formulate precise SQL queries over CSV/SQL tables instead of relying solely on vector similarity \cite{structuredrag2023}. These structured-RAG methods highlight two key advantages: (1) \emph{schema awareness}, since queries use column names and table structures directly, and (2) \emph{deterministic retrieval}, which reduces hallucinations by returning exact, verifiable values. However, most of these pipelines still rely on a static retrieval strategy (fixed chunking or always-SQL), and none dynamically adapt to table complexity or query intent. 

Beyond RAG, NL2SQL work models spreadsheet layout and headers to handle natural-language
queries over relational data \cite{jehle2025agenticnl2sqlreducecomputational}. In production
settings, domain-specific, BERT-based stacks \cite{patel2024canal,patel2024fanal} show that
smaller task-tuned models can be effective for fixed-output analytics. In parallel,
a line of \emph{tabular foundation models} has emerged early table-aware pretraining
(TaBERT, TAPAS, TABBIE) \cite{yin2020tabert,herzig2020tapas,zhang2021tabbie} and more recent
transformers (TabPFN/TabPFNv2 \cite{hollmann2025tabpfn,ye2025closerlooktabpfnv2}, TabICL
\cite{qu2025tabicl}, TabDPT \cite{ma2024tabdpt}) with strong results on supervised prediction
over single tables; surveys map the space and motivations \cite{vanbreugel2024whytfm}. However,
these lines typically assume normalized schemas or fixed targets and are not aimed at open-ended
QA over heterogeneous spreadsheets with multi-row headers, merges, and unit lines.

Our work, SQuARE, builds on these ideas but takes a different direction. Instead of committing to one retrieval method, SQuARE adapts retrieval based on the table’s structure and the question’s intent. We compute a continuous structural complexity score (e.g., header depth, merged-cell density) and use it to choose between boundary-aware semantic chunking and database querying. A lightweight LLM agent then decides, per question, whether to use semantic retrieval (FAISS), SQL, or a blend of both. This design preserves the hierarchical structure when producing embeddings and still allows for exact numeric reasoning through SQL. By adapting retrieval instead of fixing it, SQuARE reduces fragmentation, lowers hallucination rates, and provides fast, reliable QA over complex spreadsheets.

\section{Methodology}

\subsection{Overview and Problem Definition}
Given a workbook \(W\) (Excel/CSV) and a natural-language query \(q\), the system returns an answer \(a\) and the evidence \(E\) (cells, rows, or blocks) used to produce it:
\[
(a, E) \;=\; \mathrm{SQuARE}(W, q).
\]

Sheets differ in structure: some have nested headers and merged cells; others are flat with a single header row. Treating both families identically is brittle. SQuARE addresses this by (i) estimating structural complexity, (ii) building only the indices required for that sheet, and (iii) routing each query to the safest path. The complete procedure is given in Algorithm~\ref{alg:SQuARE}: it begins with complexity scoring \textsmaller{(lines~1--3)}, continues with index construction \textsmaller{(lines~4)}, proceeds to mode selection \textsmaller{(lines~5--10)}, and finally executes retrieval with checks and fallback \textsmaller{(lines~11--36)}.

\begin{algorithm}[!t]
\caption{SQuARE: Complexity-Aware Retrieval}
\label{alg:SQuARE}
\begin{algorithmic}[1]
\Require workbook $W$, query $q$, weights $(\alpha,\beta)$, thresholds $(\tau,\theta)$, token budget $T$, top-$k$
\Ensure answer $a$ with evidence, or ``abstain''

\Statex \textbf{Stage A: Score structure and build indexes}
\State $H \gets \texttt{header\_depth}(W)$
\State $M \gets \texttt{merged\_cell\_count}(W)$
\State $\mathcal{X} \gets \alpha H + \beta M$
\State $\mathcal{I}_c \gets \texttt{BuildVectorIndex}(W)$ 
\State $\texttt{flat} \gets (\mathcal{X} < \tau)$
\If{\texttt{flat}}
  \State $\mathcal{I}_s \gets \texttt{BuildDatabase}(W)$
\EndIf

\Statex \textbf{Stage B: Choose retrieval mode with an Agent}
\State $m \gets \texttt{AgentDecision}(q,\mathcal{X},\texttt{flat})$  \\
{$m \in \{\texttt{chunk},\texttt{sql}\}$}

\Statex \textbf{Stage C: Primary attempt}
\If{$m=\texttt{sql}$ \textbf{and} \texttt{flat}}
  \State $R \gets \texttt{GenerateAndRunSQL}(\mathcal{I}_s,q)$
  \If{$\texttt{QualityCheck}(R,\theta)$}
    \State \Return $\texttt{AnswerLLM}(q,R)$
  \EndIf
\EndIf
\State $C \gets \texttt{RetrieveChunks}(\mathcal{I}_c,q,k)$
\If{$\texttt{QualityCheck}(C,\theta)$}
  \State \Return $\texttt{AnswerLLM}(q,C)$
\EndIf

\Statex \textbf{Stage D: Alternate and merge (flat sheets)}
\If{\texttt{flat} \textbf{and} $m=\texttt{chunk}$}
  \State $R \gets \texttt{GenerateAndRunSQL}(\mathcal{I}_s,q)$
  \If{$\texttt{QualityCheck}(R,\theta)$}
    \State \Return $\texttt{AnswerLLM}(q,R)$
  \EndIf
\EndIf
\If{\texttt{flat}}
  \State $ctx \gets \texttt{Merge}(C,R)$
  \If{$\lVert ctx \rVert > T$}
    \State $ctx \gets \texttt{Summarize}(ctx)$
  \EndIf
  \If{$\texttt{QualityCheck}(ctx,\theta)$}
    \State \Return $\texttt{AnswerLLM}(q,ctx)$
  \EndIf
\EndIf

\State \Return ``abstain''
\end{algorithmic}
\end{algorithm}

\subsection{Adaptive Complexity Detection}

SQuARE estimates a sheet’s structural complexity from two observable quantities: header depth \(H\) and the count of merged or split cells within the header region \(M\). As computed in Algorithm~\ref{alg:SQuARE} (Stage~A, lines~1--3), the score
\[
\mathcal{X} \;=\; \alpha\,H \;+\; \beta\,M
\]
is evaluated once for the current sheet.

To avoid a single global cutoff, we apply a sheet-normalized rule. Let \(S_h\) denote the number of header cells after expanding merges (the header span), and define the merge density \(d := M / S_h\). We then classify:
\[
\text{Class}(W) \;=\;
\begin{cases}
\text{Multi-Header}, & \text{if } H \ge 2 \text{ or } d \ge \rho,\\[2pt]
\text{Flat}, & \text{otherwise,}
\end{cases}
\]
with a small fraction \(\rho \in [0.10,0.15]\) in our experiments. In Algorithm~\ref{alg:SQuARE}, the symbol \(\tau\) represents this sheet-normalized boundary via \texttt{flat} \( := (\mathcal{X} < \tau(S_h))\) (Stage~A, line~6), so the decision scales with the header span rather than a fixed global number.\footnote{Equivalently, \(\tau\) can be viewed as a function of \(S_h\);}

\begin{figure*}[!t]
  \centering
  \includegraphics[width=\textwidth]{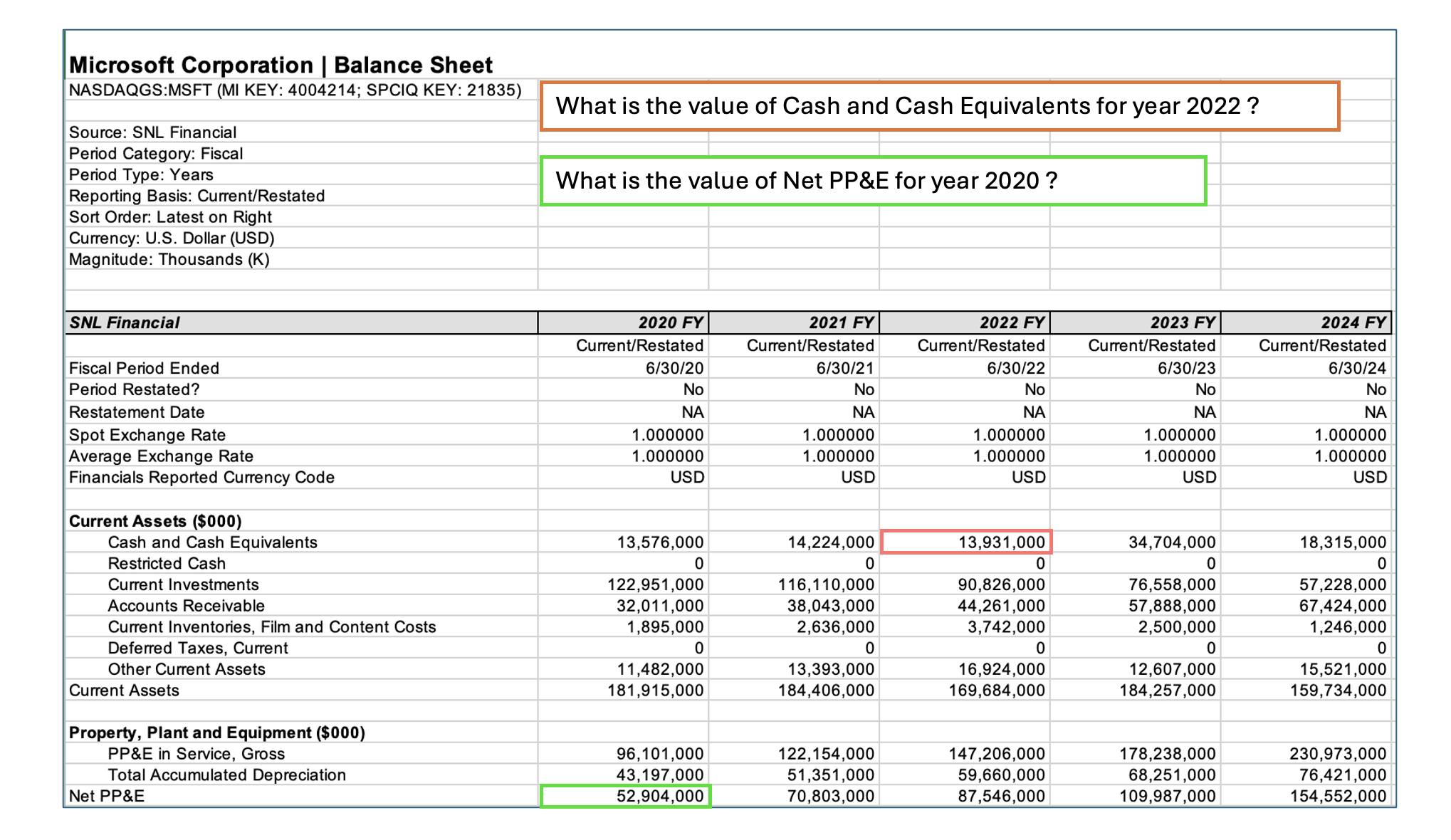}
  \caption{Example balance sheet (Microsoft, FY2020–2024). Queries such as
  \emph{Cash and Cash Equivalents (2022)} and \emph{Net PP\&E (2020)}
  require the header path, year labels, and the unit line to remain intact, motivating the Multi-Header classification and the structure-preserving retrieval path.}
  \label{fig:balance_sheet_example}
\end{figure*}

This label determines which indices are materialized next. For all sheets, the semantic index \(\mathcal{I}_c\) is built (Stage~A, line~4). For sheets classified as \emph{Flat}, the relational view \(\mathcal{I}_s\) is also constructed (Stage~A, lines~7--9). The effect is that Multi-Header sheets route to structure-preserving chunk retrieval, while Flat sheets enable both chunk retrieval and SQL over an inferred schema. See Figure~\ref{fig:balance_sheet_example} for a representative Multi-Header sheet, where header paths, year labels, and units must be preserved to answer correctly.

\subsection{Semantic Chunking with Structural Fidelity}


When a sheet is labeled \emph{Multi-Header}, SQuARE builds and uses only the semantic index \(\mathcal{I}_c\) (Stage~A, line~4). We segment at header boundaries so that each block \(B_i\) is a coherent region rather than a fixed window, and we attach explicit metadata
\[
\mathrm{meta}(B_i) \;=\; (H_i,\; Y_i,\; U_i),
\]
where \(H_i\) is the ordered header path (outer to inner labels), \(Y_i\) are the time labels present in the block, and \(U_i\) records the unit string when available.

Each block receives a brief two-sentence description \(D_i\) that states what the block contains and which years or units appear. We embed \(D_i\) to \(e_i\) and form the index entries \((e_i,B_i)\in\mathcal{I}_c\). At query time (Stage~C, lines~17--19), the algorithm retrieves a list \(C\) of the top-\(k\) \emph{blocks \(B_i\)} ranked by cosine similarity to the query embedding; we forward at most three blocks to the answer model to keep the context length stable while preserving numerical fidelity.

A quality gate (Stage~C, lines~20--22) rejects weak contexts. If the retrieved blocks pass, the system answers using only the corresponding cells. If they do not, control follows Algorithm~\ref{alg:SQuARE} to the alternate path or to merging on Flat sheets (Stage~D, lines~23--31), including summarization if the merged context exceeds the token budget.

\subsection{Structured SQL Retrieval for Flat Spreadsheets}
For Flat sheets, Algorithm~\ref{alg:SQuARE} \textsmaller{(lines 5-7)} materializes a small relational view \(\mathcal{I}_s\) then infers a schema \(\mathcal{S}=\{(c_j,t_j)\}\) with cleaned column names and types, and persists any detected units in a dedicated field. In the \texttt{sql} branch Algorithm~\ref{alg:SQuARE} (\textsmaller{(lines 11-14)}), an LLM proposes \(Q_\mathrm{SQL}\) from \((q,\mathcal{S},\text{SampleRows})\), which we execute and refine once or twice for errors or empty results. Returned rows \(R\) then serve as evidence. We reject ill-shaped outputs (for example, empty rows for scalar questions) in the subsequent quality checks.

Constrained SQL : We restrict generation to \texttt{SELECT–FROM–WHERE–GROUP BY–ORDER BY–LIMIT} over a whitelisted schema, with a strict row cap and no DDL/DML. Empty or ill-shaped results fail the same quality gate used for chunks.

\subsection{Agent-Based Intelligent Retrieval Selection}
The agent in Algorithm~\ref{alg:SQuARE} \textsmaller{(lines~9)} chooses between \texttt{chunk} and \texttt{sql} using the sheet label and cues in the question. Aggregations with explicit filters tend to route to SQL on Flat sheets. Queries that refer to layout or nested headers route to chunk retrieval. 

We apply a strict acceptance test quality check to the context returned by the chosen mode. If the primary attempt fails this test, Stage~D executes the alternate path when the sheet is Flat. If both paths produce partial confidence scores, we merge them into a single context, summarize if the token budget \(T\) is exceeded, and provide answers only if the merged context passes the same quality test. Otherwise, the system abstains. This sequence ensures that SQL is never attempted on Multi-Header sheets and that the model never answers from low-confidence contexts. We compute a single \emph{confidence score} $s_{\text{ctx}}$ to accept, switch, or merge:
for chunks, $\max_i \cos(q,e_i)$ with header/unit consistency checks; for SQL,
a combination of non-emptiness, selectivity, and schema-coverage heuristics.

\subsection{Robust Fallback and Failure Handling}
If the primary path produces weak evidence (low similarity, an empty SQL result, or an abstention), we proceed as follows,
\begin{enumerate}
  \item \textbf{Alternate retrieval.} Try the other path if available (Flat sheets only), preserving the same quality checks.
  \item \textbf{Context merging.} Combine the top chunks \(C\) with the SQL result \(R\) (\texttt{Merge} in Algorithm~\ref{alg:SQuARE} \textsmaller{line~28)}. If the merged context exceeds the budget \(T\), we summarize it \textsmaller{(line~30)}.
  \item \textbf{Answer or abstain.} If the merged context passes the quality check, we answer; otherwise, we abstain.
\end{enumerate}

\subsection{Scalability and Efficiency}
We build \(\mathcal{I}_c\) once per sheet and materialize \(\mathcal{I}_s\) only when needed, Algorithm~\ref{alg:SQuARE} (\textsmaller{line~7)}). Retrieval results are cached (query\(\to\)chunk IDs; SQL string\(\to\)rows). Components are swappable without changing control flow: models, embeddings, vector engines, and databases.

\subsection{Hyperparameter Settings}
We tuned the required parameters on a held-out development split. We use \(\alpha{=}0.6\) and \(\beta{=}0.4\). The sheet classifier applies a data-calibrated threshold \(\tau^\star\). At inference, we forward at most \(k{=}3\) chunks from \(\mathcal{I}_c\) and use a single confidence threshold \(\theta\) selected on the development split. No dataset-specific tuning was performed.

\section{Datasets}

To evaluate \textit{SQuARE} across the structural spectrum, we use three categories: (i) complex multi-header corporate spreadsheets, (ii) a complex merged World Bank workbook with heterogeneous sheets, and (iii) flat, single-header public tables. Table~\ref{tab:datasets} summarizes the sources and QA volumes.

\begin{table*}[t]
\centering
\caption{Evaluation datasets and QA configuration.}
\label{tab:datasets}
\begin{tabular}{l l l}
\hline
\textbf{Category} & \textbf{Sources} & \textbf{QA Pairs / Tiers} \\
\hline
Complex Multi‐Header Financial Spreadsheets &
\shortstack[l]{Microsoft, Meta, Alphabet, Netflix, Tesla, \\ Adobe, SAP, Nvidia, Amazon, Dell Technologies} &
40 QA per workbook, 400 QA total \\
\hline
Complex Merged World Bank Benchmark &
\shortstack[l]{Gender Statistics + World\\Development Indicators merged\\into one workbook} &
50 QA total \\
\hline
Simple Structured Tables &
\shortstack[l]{Health \& Nutrition; Public Sector Debt;\\Global Economic Prospects; Energy Consumption;\\Education Attainment} &
\shortstack[l]{Easy / Medium / Hard tiers,\\30 QA each tier per dataset, 450 QA total} \\
\hline
\end{tabular}
\end{table*}

\subsection{Dataset Types}
\subsubsection{Complex Multi-Header Financial Spreadsheets}
We collected quarterly or annual balance sheets from ten large issuers (Microsoft, Meta, Alphabet, Netflix, Tesla, Adobe, SAP, Nvidia, Amazon, and Dell Technologies). These files exhibit multi-row nested headers, frequent merged cells, section resets, and unit lines between headers and values. For each workbook, we authored 40 QA pairs (400 total) that require resolving header paths, unit strings, and year columns. A representative query is, \emph{“What percentage did Net PP\&E increase from 2020 to 2023?”} See Fig.~\ref{fig:balance_sheet_example} for a balance-sheet example.

\subsubsection{Merged World Bank Benchmark}
We merged World Bank Gender Statistics and World Development Indicators into a single workbook with non-uniform layouts. The set contains 50 QA pairs that force the alignment of demographic and economic indicators across years.

\subsubsection{Simple Structured CSV/Excel Tables}
We selected five flat, single-header datasets (Health and Nutrition, Public Sector Debt, Global Economic Prospects, Energy Consumption, Education Attainment). For each dataset, we created 90 QA pairs, stratified as 30 Easy, 30 Medium, and 30 Hard; this yields a total of 450 QA pairs overall. Easy items involve direct lookups, Medium items add filters and simple aggregates, and Hard items require multi column predicates, groupings, or comparisons across entities/years.

\begin{table}[t]
  \small
  \centering
  \caption{Difficulty level examples for flat tables.}
  \label{tab:qa_difficulty}
  \setlength{\tabcolsep}{6pt}              
  \setlength{\extrarowheight}{2pt}         
  \renewcommand{\arraystretch}{1.22}       
  \begin{tabularx}{\columnwidth}{|l|>{\raggedright\arraybackslash}X|}
    \hline
    \textbf{Difficulty} & \textbf{Example question} \\
    \hline
    Easy   & What was the fertility rate of Country~X in 2018? \\
    \hline
    Medium & What was the mean public debt as \%~of GDP for Country~Y from 2015 to 2020? \\
    \hline
    Hard   & Which countries had GDP growth greater than 3\% in 2019? \\
    \hline
  \end{tabularx}
\end{table}

\subsection{Question Taxonomy and Evaluation Protocol}
Questions are grouped by retrieval pathway: header driven (multi-header and merged WB), SQL-driven (flat tables with tiers), and hybrid (merge step when the primary mode yields low confidence). We report exact-match accuracy for numeric and categorical answers, as well as retrieval recall separately for chunk retrieval and SQL results, isolating retrieval quality from answer generation.

\textbf{Difficulty rubric} \emph{Easy}: direct lookup; \emph{Medium}: one aggregate or range filter;
\emph{Hard}: multi-predicate filters and/or comparative aggregates across entities/years. 

\section{Evaluation and Results}

We evaluate \textit{SQuARE} across three settings: multi header corporate spreadsheets, a complex merged World Bank workbook, and flat public tables -- and compare against strong baselines, including ChatGPT-4o.

\subsection{Evaluation Protocol}

We report exact match accuracy for numeric and categorical answers. Exact match avoids semantic scoring ambiguity (e.g., ``1 billion'' vs. ``1{,}000 million'') where metrics such as BERTScore~\cite{BERTScore} are not faithful to numeric correctness. To isolate retrieval quality from answer generation, we additionally report retrieval recall (R@$k$).
For ChatGPT-4o, we report a tool-free, out of the box baseline (no external retrieval or SQL tools)
to reflect default assistant behavior.

\textbf{Models and controls.}
We instantiate SQuARE with Gemma~3:12B~\cite{team2024gemma} and Llama~3.2:11B~\cite{llama3.1} (the same model is used throughout routing and answering). For the ChatGPT-4o baseline, we issue the same questions via the publicly available ChatGPT-4o web interface, with no tool use, browsing, or external retrieval, to reflect an out-of-the-box setting. Across all runs, we use temperature~0 and a fixed top-$k$ for retrieval. All outputs were manually verified due to the prevalence of numeric answers.

\subsection{Results on Multi-Header and Merged Benchmarks}

\begin{figure}[!t]
  \centering
  \includegraphics[width=\columnwidth]{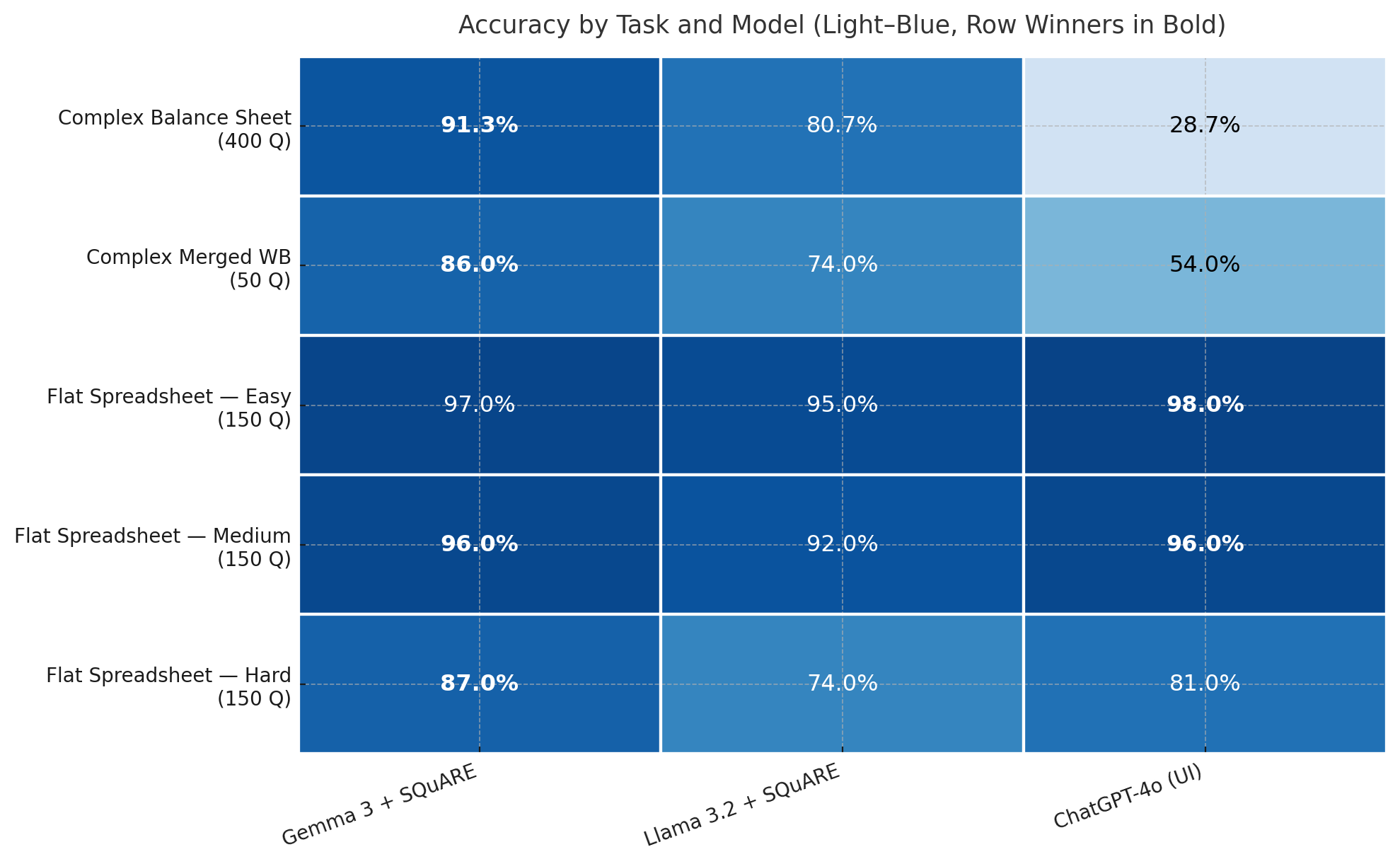}
  \caption{Accuracy by task and model. The router chooses chunk vs constrained SQL; the confidence gate switches or merges when context is weak. Darker cells indicate higher accuracy; bold marks the row winner.}
  \label{fig:acc_confusion_grid_colored}
\end{figure}

Figure~\ref{fig:acc_confusion_grid_colored} summarizes accuracy per task and model. On multi header corporate balance sheets, SQuARE(Gemma) attains \textbf{91.3\%}, outperforming SQuARE(Llama) (80.7\%) and ChatGPT-4o (28.7\%). On the merged World Bank workbook, SQuARE(Gemma) reaches \textbf{86.0\%}, ahead of SQuARE(Llama) (74.0\%) and ChatGPT-4o (54.0\%). These gains are consistent with the design goal of preserving header paths and units when the layout is complex and presenting verifiable rows as evidence.

\subsection{Results on Flat, Single Header Tables}

On five flat datasets (Health, Public Debt, GEP, Energy, Education; 450 QA total with 30 Easy/30 Medium/30 Hard per dataset), SQuARE’s dual indexing (vector + SQL) and routing deliver strong performance. SQuARE(Gemma) achieves \textbf{93.3\%} overall, with \textbf{87\%} on the Hard tier; ChatGPT-4o is competitive on Easy/Medium but trails on Hard (81\%), where exact filters, groupings, and multi-column predicates favor SQL-grounded retrieval. SQuARE(Llama) is consistently behind Gemma but benefits from the same routing.

\subsection{Retrieval Recall@\texorpdfstring{$k$}{k}}
\label{sec:retrieval_recall}

Answer accuracy conflates retrieval failure with reasoning errors. We therefore report R@$k$ for each path. Let $\mathcal{E}$ be the set of ground-truth evidence units (chunks that contain the answer for vector retrieval or rows returned by the SQL). For query $q$, if the system returns the top $k$ retrieved evidence $\mathcal{R}_k$, then
\[
\text{R@}k(q)=\frac{|\mathcal{R}_k \cap \mathcal{E}|}{|\mathcal{E}|},
\]
averaged over all queries. Because SQuARE forwards at most three chunks or one SQL result set to the LLM, we fix $k=3$ for vector retrieval and $k=1$ for SQL.

\begin{table}[t]
  \small
  \centering
  \caption{Retrieval recall@\textit{k}. Higher is better.}
  \label{tab:recall_results}
  \setlength{\tabcolsep}{8pt}        
  \renewcommand{\arraystretch}{1.35} 
  \begin{tabularx}{\columnwidth}{|>{\raggedright\arraybackslash}X|c|c|c|}
    \hline
    \textbf{Dataset} & \textbf{FAISS R@3} & \textbf{SQL R@1} & \textbf{Merged R@3} \\ \hline
    10 Multi-Header Balance Sheets (400 Q) & 0.88 & --   & --   \\ \hline
    Merged World Bank Workbook (50 Q)      & 0.86 & --   & --   \\ \hline
    Flat Tables - Easy                     & 0.87 & 0.95 & 0.89 \\ \hline
    Flat Tables - Medium                   & 0.85 & 0.92 & 0.88 \\ \hline
    Flat Tables - Hard                     & 0.81 & 0.90 & 0.84 \\ \hline
  \end{tabularx}
\end{table}

\textbf{Observation} For complex sheets, the chunk retriever surfaces $\ge\!0.86$ of the necessary evidence within three chunks. On flat tables, SQL achieves near-perfect R@1, while the vector path contributes marginal cases in the Hard tier (Table~\ref{tab:recall_results}).

\subsection{Ablation: Role of Fallback, Routing, and SQL}

We study three ablations against the full system: (i) \emph{No Fallback} (disable alternate/merge), (ii) \emph{Chunk-only} (no SQL on flat tables), and (iii) \emph{SQL-only} (use SQL on flat tables, not applicable on multi-header).

\begin{table}[t]
  \small
  \centering
  \caption{Ablation study: overall accuracy by variant.}
  \label{tab:ablation}
  \setlength{\tabcolsep}{7pt}        
  \renewcommand{\arraystretch}{1.2}  
  \begin{tabularx}{\columnwidth}{|>{\raggedright\arraybackslash}X|c|c|c|}
    \hline
    \textbf{Variant} &
    \textbf{\shortstack{Multi-Header\\(400 Q)}} &
    \textbf{\shortstack{Merged WB\\(50 Q)}} &
    \textbf{\shortstack{Flat\\(450 Q)}} \\
    \hline
    Full SQuARE          & \textbf{91.3\%} & \textbf{86.0\%} & \textbf{93.3\%} \\
    \hline
    No Fallback          & 89.0\%          & 80.0\%          & 90.0\%          \\
    \hline
    Chunk-only           & 89.0\%          & 80.0\%          & 75.5\%          \\
    \hline
    SQL-only (Flat only) & --              & --              & 90.0\%          \\
    \hline
  \end{tabularx}
\end{table}

Removing fallback costs 4-6 points across settings, confirming that confidence-guided switching/merging is material. On flat tables, \emph{Chunk-only} underperforms \emph{SQL-only}, consistent with the advantage of deterministic filters and aggregates; the full system recovers the remaining gap via routing and merging.
\subsection{LLM Invocation Budget}

\begin{figure}[!t]
  \centering
  \includegraphics[width=\columnwidth]{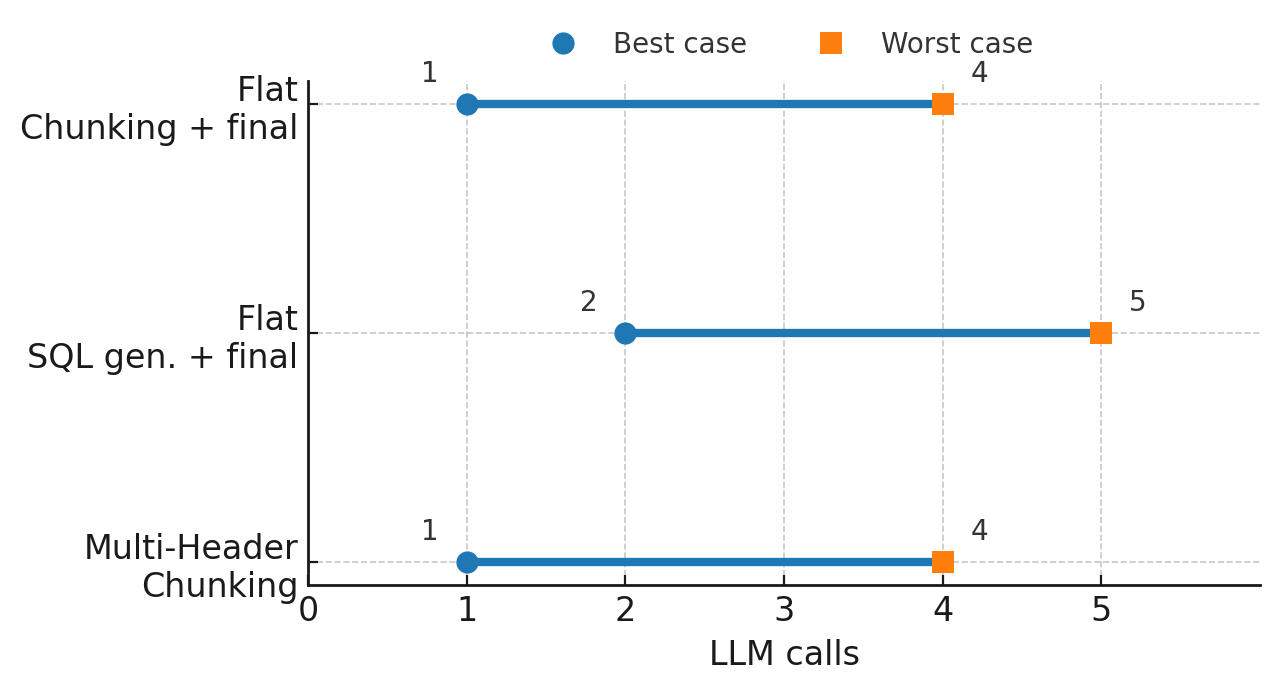} 
  \caption{LLM Invocation Budget per Query}
  \label{fig:invocation_budget}
\end{figure}

We distinguish between offline (indexing) and online (query time) LLM calls. Offline, chunk descriptions are generated once in a batched step. Online, per-query LLM calls depend on spreadsheet type and retrieval path, as summarized in figure ~\ref{fig:invocation_budget}.

In practice, most queries complete within the best case budget, ensuring low latency and predictable costs. Across 400 multi-header queries, SQuARE chose chunking 100\% of the time (fallback $<$ 8\%). For flat tables, the agent selected chunking 35\% and SQL 65\%, with an overall fallback rate of $\approx$ 10\%.

\subsection{Benchmarking against ChatGPT-4o \texorpdfstring{\cite{gpt4o}}{[gpt4o]}}

Comparing SQuARE directly against ChatGPT-4o highlights significant performance differences, especially on structurally intricate and context rich queries.

Gemma based SQuARE consistently outperformed ChatGPT-4o, particularly in challenging multi header contexts. For instance, complex balance sheet questions achieved over a threefold improvement (91.3\% vs. 28.7\%). On simpler structured queries, SQuARE maintained competitive or better performance, notably surpassing ChatGPT-4o at high difficulty tiers (87\% vs. 81\%).

Analysis of errors revealed ChatGPT-4o’s frequent confusion due to multiple relevant numerical entries, unit mismatches, or an inability to accurately reference specific table rows and columns. Conversely, SQuARE’s semantic chunking and structured querying with fallback effectively mitigated these errors.

\subsection{Efficiency and Cost Analysis}

On our reference setup (quantized models running on a T4 GPU with 15 GB VRAM), SQuARE’s end-to-end latency was close to that of ChatGPT-4o and varied with the sheet structure and the retrieval path chosen in Algorithm~\ref{alg:SQuARE} \cite{GoogleColab}. On complex financial spreadsheets, SQuARE was consistently a little slower with the same prompts. On flat, single-header files, the gap was small and often narrowed further.

The difference aligns with the invocation budget: the chunk path issues one answer call after retrieval, whereas the SQL path adds a brief SQL-generation step before answering.

SQuARE runs comfortably on modest hardware. The experiments above used quantized models on T4 GPUs; moving to unquantized models or newer GPUs, such as the A100 or H100, further reduces latency \cite{d2lGPUarch}. In practice, the framework trades a small amount of compute time for a consistent gain in answer fidelity on structurally complex spreadsheets, without requiring specialized infrastructure.

\section{Limitations and Future Work}

SQuARE targets spreadsheet QA with an emphasis on numeric fidelity and evidence. The system is effective within that scope but also exposes a few limitations that suggest clear next steps.

\textbf{Router learning and calibration} The agentic router is prompt based. A lightweight learned router with uncertainty estimates could reduce fallbacks and sharpen confidence checks. A cost aware objective that balances accuracy, context length, and latency is a natural extension.

\textbf{Layout variability and formats} Experiments focused on well structured spreadsheets. Handling OCR’d tables and document style layouts will require integrating table detectors and layout parsers before applying the same retrieval backbone. Extending SQuARE beyond spreadsheets to document-style inputs may follow a path similar to \emph{MultiFinRAG's} \cite{Gondhalekar2025MultiFinRAG} multimodal extraction and tiered fusion while retaining SQuARE’s SQL/structure routing in tabular regions.

\textbf{SQL robustness and join scope} SQL generation can be fragile under schema drift. We will add schema alignment, column–header aliasing, and safe join discovery. Beyond single sheets, we plan to support multi sheet and cross workbook queries with explicit, verifiable joins while maintaining the evidence first interface.

\textbf{Evaluation coverage} We used manual grading and compared it against ChatGPT–4o in a public, tool free setting. A broader baseline set that includes tool augmented assistants, perturbation tests for robustness, and the public release of de identified QA pairs (where licensing permits) is a priority.

\textbf{Model recency} While this manuscript was being finalized, newer general purpose models GPT–5\cite{openai2025gpt5-intro}, GPT-oss\cite{openai2025gptoss120bgptoss20bmodel} became available. We did not have time for extensive testing; early smoke tests on a small slice were promising. A controlled, like for like comparison under identical retrieval conditions and constraints is future work.

\textbf{Tabular foundation models (TFMs)} Recent surveys argue for TFMs as a distinct modeling track for tables \cite{vanbreugel2024whytfm,lu2025llmfortables}, and early models show a strong capacity for structure and cross table reasoning \cite{ma2024tabdpt,zhang2024xtformer,qu2025tabicl}. Within SQuARE, TFMs would serve as drop in encoders for structure and units, replacing only the embedding and structure components while preserving the RAG+SQL orchestration; component swaps then allow targeted fine tuning as TFMs mature.

\section{Conclusion}

SQuARE is an agentic RAG+SQL framework that, for each sheet and each question, selects between structure preserving chunk retrieval and schema aware SQL, and returns answers with the exact spreadsheet cells or rows as evidence. Across approximately 900 QA pairs spanning complex balance sheets, a complex merged World Bank workbook, and flat spreadsheets with difficulty tiers, SQuARE consistently outperformed a widely available ChatGPT-4o baseline under identical prompts and without external tools, with the largest gains on structurally complex files. Ablations and retrieval level recall isolate the lift; the complexity gate prevents brittle choices, chunking strategy preserves header paths and units, and the SQL path provides deterministic filtering and aggregation on flat tables. The system is practical to deploy, with embeddings, vector stores, and databases being swappable. Lazy indexing, along with caching, keeps runtime predictable on modest hardware, and it yields answers that are accurate and directly traceable to the underlying spreadsheets.

\bibliographystyle{IEEEtran}
\bibliography{references}  

\end{document}